\documentclass[sigconf]{acmart}

\def\BibTeX{{\rm B\kern-.05em{\sc i\kern-.025em b}\kern-.08emT\kern-.1667em\lower.7ex\hbox{E}\kern-.125emX}}

\renewcommand\footnotetextcopyrightpermission[1]{} 
\usepackage{amsmath}
\usepackage{amsfonts}
\usepackage{graphicx}
\usepackage{textcomp}
\usepackage{xcolor}
\usepackage{tcolorbox}
\usepackage{bbold}
\usepackage{balance}
\usepackage[marginal,para]{footmisc}
\usepackage{soul}
\definecolor{shadecolor}{rgb}{0.9425,0.9425,0.9425}

\definecolor{secolor}{rgb}{0.4,0.4,0.4}

\usepackage{colortbl}
\definecolor{Gray}{gray}{0.9}

\def\BibTeX{{\rm B\kern-.05em{\sc i\kern-.025em b}\kern-.08em
    T\kern-.1667em\lower.7ex\hbox{E}\kern-.125emX}}

\usepackage{adjustbox}
\usepackage{array}
\usepackage{booktabs}
\usepackage{multirow}
\newcolumntype{R}[2]{%
    >{\adjustbox{angle=#1,lap=\width-(#2)}\bgroup}%
    l%
    <{\egroup}%
}
\newcommand*\rot{\multicolumn{1}{R{45}{1em}}}%
\usepackage{bbding}
\usepackage{pifont}
\newcommand{\cmark}{\color{green} \ding{51}}%
\newcommand{\xmark}{\color{red} \ding{55}}%

\usepackage[skip=0.5\baselineskip,font=small]{caption}

\usepackage{enumitem}
\setlist{leftmargin=3mm}

\begin{document}
\pagestyle{plain}
\pagenumbering{gobble}

\title{
\vspace{-22pt} 
\underline{\textit{You Only Search Once}}: On Lightweight Differentiable Architecture Search for Resource-Constrained Embedded Platforms 
\vspace{-4pt}
}
\author{Xiangzhong Luo$^1$, Di Liu$^2$, Hao Kong$^1$, Shuo Huai$^1$, Hui Chen$^1$, and Weichen Liu$^{1*}$}
\author{$^1$School of Computer Science and Engineering, Nanyang Technological University, Singapore}
\author{$^2$HP-NTU Digital Manufacturing Corporate Lab, Nanyang Technological University, Singapore}
\thanks{
$^{*}$Corresponding author: Weichen Liu (liu@ntu.edu.sg,xiangzho001@e.ntu.edu.sg).

This work is partially supported by the Ministry of Education, Singapore, under its Academic Research Fund Tier 2 (MOE2019-T2-1-071) and Tier 1 (MOE2019-T1-001-072), and partially supported by Nanyang Technological University, Singapore, under its NAP (M4082282) and SUG (M4082087).}

\begin{abstract}
Benefiting from the search efficiency, differentiable neural architecture search (NAS) has evolved as the most dominant alternative to automatically design competitive deep neural networks (DNNs). We note that DNNs must be executed under strictly hard performance constraints in real-world scenarios, for example, the runtime latency on autonomous vehicles. However, to obtain the architecture that meets the given performance constraint, previous hardware-aware differentiable NAS methods have to repeat a plethora of search runs to manually tune the hyper-parameters by trial and error, and thus the total design cost increases proportionally. To resolve this, we introduce a lightweight hardware-aware differentiable NAS framework dubbed LightNAS, striving to find the required architecture that satisfies various performance constraints through a one-time search (i.e., \underline{\textit{you only search once}}). Extensive experiments are conducted to show the superiority of LightNAS over previous state-of-the-art methods. Related codes will be released at \url{https://github.com/stepbuystep/LightNAS}.
\end{abstract}

\maketitle

\section{Introduction}
\label{sec:introduction}

Deep neural networks (DNNs) are becoming ubiquitous across a plethora of intelligent embedded applications such as virtual reality (VR) \cite{zhang2021f} and object detection/tracking \cite{zhang2019skynet}, enabling entirely new on-device experiences \cite{wu2019machine}. Nonetheless, given that the network design space is tremendously large \cite{cai2018proxylessnas, wu2019fbnet}, manually designing competitive DNNs requires considerable human efforts to determine the optimal network configuration. To address this, neural architecture search (NAS) \cite{zoph2016neural} has recently flourished, which is dedicated to automating the design of top-performing DNNs. In the literature, NAS studies can be mainly divided into reinforcement learning \cite{zoph2016neural}, evolution \cite{real2019regularized}, and gradient-based \cite{liu2018darts} (a.k.a., differentiable) categories. However, both reinforcement learning and evolution-based NAS approaches suffer from prohibitive search overheads (e.g., over 2,000 GPU days \cite{zoph2016neural} and 3,150 GPU days \cite{real2019regularized}), whereas the differentiable counterpart is of great search efficiency that dramatically reduces the search cost by multiple orders of magnitude (e.g., 1 GPU day \cite{liu2018darts}).

Despite the significant progress achieved so far, the early differentiable NAS approaches \cite{liu2018darts, xie2018snas} are hardware-agnostic since they merely focus on searching for competitive architectures in terms of the accuracy, regardless of other critical performance metrics such as the on-device latency and energy, which are of paramount importance for AI-empowered applications, especially on resource-constrained embedded platforms \cite{wu2019machine}. Among these, \cite{xie2018snas} strives to design lightweight DNNs according to the number of floating-point operations (FLOPs), but the number of FLOPs is an inaccurate proxy, which cannot accurately reflect the actual latency and energy consumption on target hardware (see Figure~\ref{fig:latency-vs-flops}). To this end, several hardware-aware differentiable NAS methods are then proposed \cite{cai2018proxylessnas, wu2019fbnet, vahdat2020unas}, which incorporate the on-device latency into the search objective as soft constraints to penalize the architecture candidate with high latency, thereby being able to generate hardware-friendly architecture solutions with low inference latency.

But even so, we, in practice, should consider not only the \underline{\textit{explicit}} search cost -- the time required to run one single search, but also the \underline{\textit{implicit}} search cost -- the time required for manual hyper-parameter tuning in order to find the desired architecture. For instance, in real-world scenarios like autonomous vehicles, DNNs must be executed under strictly hard latency constraints. Unfortunately, to obtain the architecture with competitive accuracy while satisfying the given latency constraint, previous hardware-aware differentiable NAS methods \cite{cai2018proxylessnas, wu2019fbnet, vahdat2020unas} have to perform multiple search runs to manually tune the hyper-parameters (see $\lambda$ in Eq~(\ref{eq:proxylessnas-objective})) by trial and error. As a result, the total design cost increases proportionally (empirically by $\times10$ times). Meanwhile, during the search process \cite{cai2018proxylessnas, wu2019fbnet, vahdat2020unas}, multiple sub-networks (paths) need to be optimized at the same time (see Table~\ref{tab:comparisons}), thereby leading to non-trivial memory overheads. And even worse, the above multi-path paradigm introduces considerable inconsistency between search and evaluation since the stand-alone architecture at the evaluation stage is a single-path sub-network \cite{chu2021fairnas}. As such, it is natural to ask the following question:
\begin{tcolorbox}[colback=shadecolor, boxrule=0.425mm,left=2.5pt,right=2.5pt,top=2pt,bottom=2pt,
                grow to left by=-2pt,
                grow to right by=-2pt,
                enlarge top by=-3pt,
                enlarge bottom by=-3pt]
\textit{Is it possible to find the required architecture that strictly satisfies the given performance constraint through a one-time architecture search in both differentiable and lightweight manners?}
\end{tcolorbox}

\begin{figure}[t]
    \begin{center}
        \includegraphics[width=1.0\columnwidth]{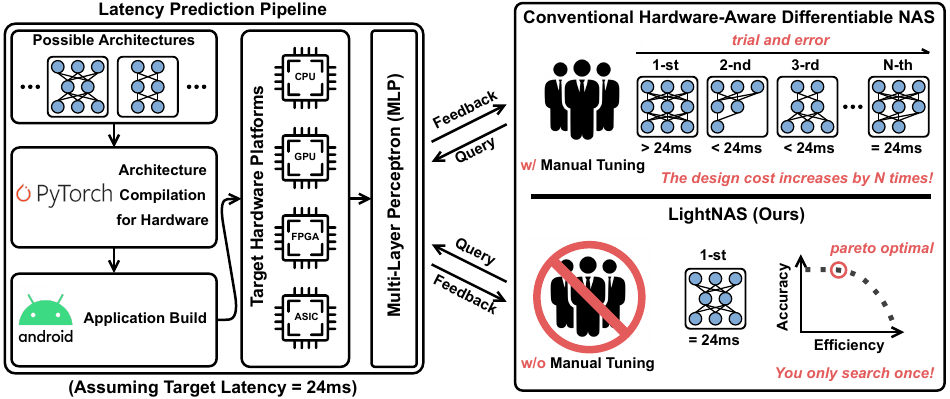}
    \end{center}
    \vspace{-2pt}
    \caption{An intuitive illustration of the proposed LightNAS method.}
    \label{fig:overview}
\end{figure}

To answer the question outlined above, we propose a lightweight hardware-aware differentiable NAS method dubbed LightNAS, in which we primarily focus on the most dominant performance constraint, i.e., latency (see Figure~\ref{fig:overview}). LightNAS is compared against previous NAS methods in Table~\ref{tab:comparisons}. Our contributions are as follows:
\begin{itemize}
    \item {
        We introduce an accurate yet efficient predictor to estimate the latency so as to avoid the tedious on-device measurements, which is also generalizable to other hardware metrics.
       \textbf{(see Section~\ref{sec:latency-prediction})}
    }
    \item {
        We propose a lightweight differentiable search method to reduce the optimization complexity to the single-path level, thereby effectively resolving the memory bottleneck.
        \textbf{(see Section~\ref{sec:lightweight-differentiable-architecture-search})}
    }
    \item{
        We incorporate the latency predictor into LightNAS to achieve hardware-aware architecture search. Instead of manually tuning the hyper-parameters to guarantee the given latency constraint, LightNAS can automatically learn the optimal hyper-parameter configuration during the search phase, thus being able to find the desired architecture through a one-time search.
        \textbf{(see Section~\ref{sec:hardware-aware-differentiable-architecture-search})}
    }
    \item{
        Extensive experiments show that LightNAS can effectively and efficiently find the architecture that meets the specified latency constraint through a one-time search, surpassing previous methods in terms of both accuracy and efficiency.
        \textbf{(see Section~\ref{sec:experiments})}
    }
\end{itemize}

\section{Background}
\label{sec:preliminaries-and-motivations}
In this section, we first introduce the preliminaries on differentiable NAS \cite{liu2018darts}, and then present the motivations of this paper.

\vspace{-4pt}
\subsection{Preliminaries on Differentiable NAS}
\label{sec:preliminaries-on-differentiable-nas}

We begin with the operator space denoted as $\mathcal{O}=\{o_k\}_{k=1}^{K}$, in which each element $o_k$ represents an operator candidate. Following the weight-sharing NAS paradigm \cite{liu2018darts}, an over-parameterized network is constructed, namely supernet, where each layer is composed of $K$ operator candidates lied in $\mathcal{O}$. To relax the discrete network design space to become continuous, operators in the supernet are assigned with a set of architecture parameters $\alpha \in \mathbb{R}^{L \times K}$, where $L$ corresponds to the number of layers in the supernet. The structure of the supernet is visualized in Figure~\ref{fig:search-space}. Therefore, we are able to formulate the output of the supernet $F(x)$ as follows:
\begin{equation}
    \small
    \setlength\abovedisplayskip{4pt}
    \setlength\belowdisplayskip{4pt}
    F(x) = \sum\nolimits_{l=1}^L\sum\nolimits_{k=1}^{K} \left( \frac{\exp(\alpha_{l}^{k})}{\sum\nolimits_{k'=1}^{K}\exp(\alpha_{l}^{k'})} \cdot o_{k}(x_l) \right), \,\, s.t., \,\, o_{k} \in \mathcal{O}
    \label{eq:softmax-relax}
\end{equation}
where $x_l$ is the input of $l$-th layer, and $x$ is the initial input. Due to the continuous relaxation, both network weights $w$ and architecture parameters $\alpha$ can be optimized with gradient descent \cite{liu2018darts}, or more specifically, a bi-level optimization strategy is applied, including a training phase to optimize $w$ and a validation phase to optimize $\alpha$:
\begin{equation}
    \small
    \setlength\abovedisplayskip{4pt}
    \setlength\belowdisplayskip{4pt}
    \begin{aligned}
        &\mathop{\mathrm{minimize}}_{\alpha} \,\, \mathcal{L}_{valid}(w^*(\alpha), \alpha) \\
        &s.t., \,\, w^*(\alpha)=\mathrm{arg\,min}_{w} \,\, \mathcal{L}_{train}(w, \alpha)
    \end{aligned}
    \label{eq:darts-objective}
\end{equation}
where $\mathcal{L}_{train}(\cdot)$ and $\mathcal{L}_{valid}(\cdot)$ are the loss functions accumulated on the training and validation datasets, respectively. Subsequently, once the architecture search process terminates, we can determine the searched architecture by reserving the strongest operator for each layer while other operators are discarded, where the operator strength is defined as $\exp(\alpha_{l}^{k})/\sum_{k'=1}^{K} \exp(\alpha_{l}^{k'})$. For more technical details about differentiable NAS, you may refer to DARTS \cite{liu2018darts}.

\begin{figure}[t]
    \begin{center}
        \includegraphics[width=1.0\columnwidth]{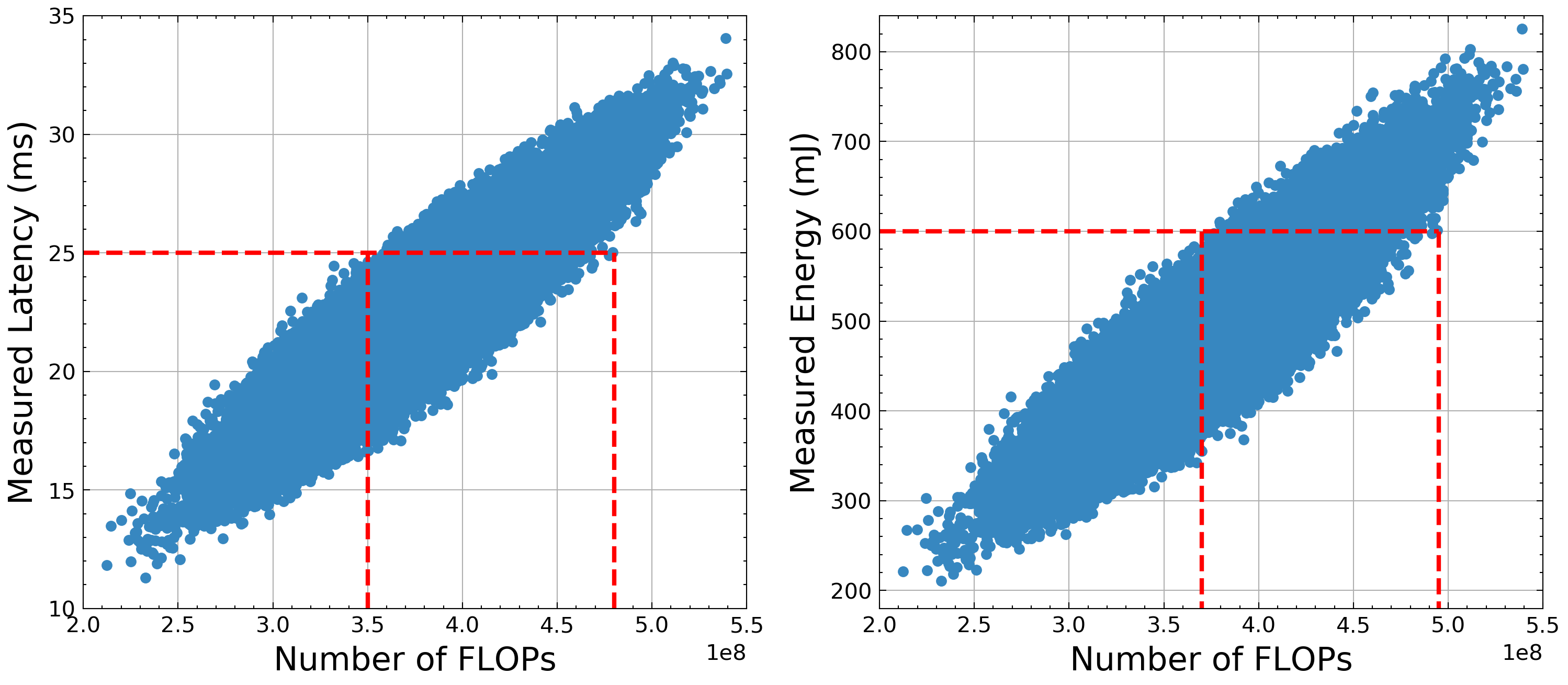}
    \end{center}
    \vspace{-2pt}
    \caption{Relationships between the number of FLOPs and the actual latency (\textit{Left}) / energy (\textit{Right}) measured on Nvidia Jetson AGX Xavier.}
    \label{fig:latency-vs-flops}
\end{figure}

\vspace{-4pt}
\subsection{Motivations}
\label{sec:motivations}

The objective defined in Eq~(\ref{eq:darts-objective}), however, focuses on the accuracy-only optimization, regardless of other critical performance metrics like the latency and energy on target hardware. As a result, it derives the architecture with competitive accuracy, which comes at the cost of extremely high computational complexity, and thus cannot be deployed on resource-limited embedded systems \cite{wu2019machine}. To tackle this issue, previous NAS methods \cite{xie2018snas, tan2019efficientnet} exploit the multi-objective optimization scheme to achieve trade-offs between accuracy and efficiency, where they use hardware-agnostic metrics like FLOPs to denote the network efficiency. Unfortunately, the number of FLOPs does not always reflect the on-device latency and energy as shown in Figure~\ref{fig:latency-vs-flops}, where we find that architectures with the same latency or energy could greatly differ regarding the number of FLOPs. 

\begin{figure}[t]
    \begin{center}
        \includegraphics[width=1.0\columnwidth]{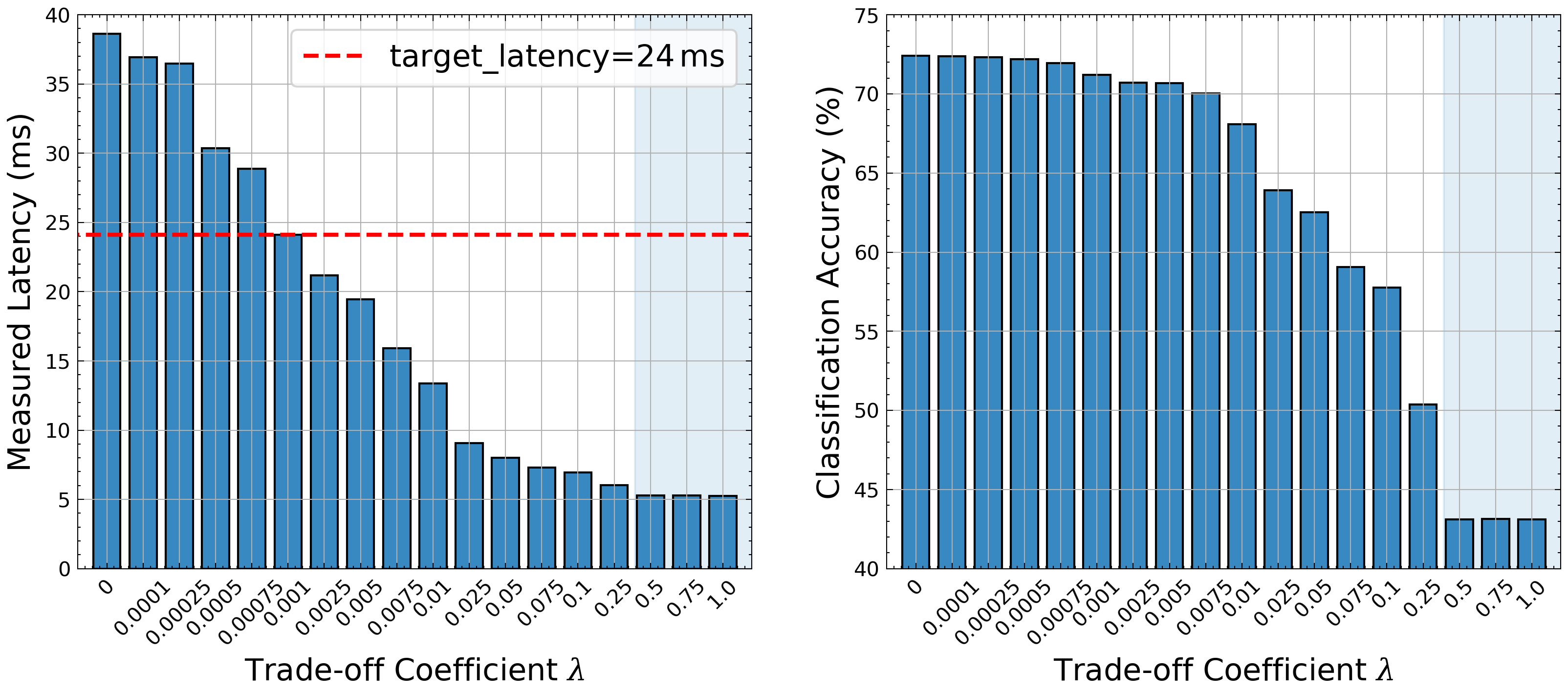}
    \end{center}
    \vspace{-2pt}
    \caption{Illustration of the search results under $\lambda \in [0,1]$ regarding the latency on Xavier (\textit{Left}) and the accuracy on ImageNet (\textit{Right}). $\lambda > 0.25$ leads to the architectures that only consist of \textit{SkipConnect}.}
    \vspace{-1pt}
    \label{fig:tradeoff}
\end{figure}

Subsequently, several hardware-aware differentiable NAS works are proposed \cite{cai2018proxylessnas, wu2019fbnet, vahdat2020unas, kim2021mdarts}, i.e., they integrate the on-device latency into the optimization objective to penalize the architecture with high latency, which can be formally expressed as follows:
\begin{equation}
    \small
    \setlength\abovedisplayskip{3.25pt}
    \setlength\belowdisplayskip{1.5pt}
    \mathop{\mathrm{minimize}}_{\alpha} \,\, \mathcal{L}_{valid}(w^*(\alpha), \alpha) + \lambda \cdot LAT(\alpha)
    \label{eq:proxylessnas-objective}
\end{equation}
where $LAT(\alpha)$ denotes the latency of the architecture encoded by $\alpha$. $\lambda \geq 0$ is a constant to control the trade-off magnitude between accuracy and latency. In fact, the above optimization objective is able to derive hardware-friendly architectures with both high accuracy and low latency, \underline{\textit{but only if a suitable $\lambda$ is applied}}. The intuition behind this is that, if $\lambda$ is too small, the latency penalty term $LAT(\alpha)$ will be effectively ignored. In contrast, if $\lambda$ is too large, we will end up with the architecture that has extremely low latency on target hardware but sub-optimal accuracy on target task. Meanwhile, in real-world scenarios like autonomous vehicles, DNNs must be executed under strictly hard latency constraints \cite{wu2019machine}. Thus, to find the architecture that satisfies the specified latency constraint, we have to perform a hyper-parameter sweep to manually tune $\lambda$ by trial and error.

Nonetheless, the above manual hyper-parameter sweep requires us to run multiple searches (empirically 10), thereby increasing the total design cost by $\times10$ times. To illustrate this point, we present a motivational experiment, in which we take FBNet \cite{wu2019fbnet} as the search engine and perform a series of search experiments under different settings of $\lambda \in [0,1]$. Then, after the search process is finished, we train the searched architecture from scratch on ImageNet for 50 epochs to quickly evaluate the accuracy. Meanwhile, we measure the corresponding latency on Nvidia Jetson AGX Xavier. As seen in Figure~\ref{fig:tradeoff}, $\lambda$ can effectively control the trade-off magnitude between accuracy and latency, but it is quite difficult to tune. For example, to obtain the architecture with the required latency of 24\,ms, we should set $\lambda$ to 0.001. But, next time if we require an architecture with the latency of 26\,ms, we need to manually tune $\lambda$ within the range of $[0.00075,0.001]$ as shown in Figure~\ref{fig:tradeoff}, inevitably leading to a plethora of trial-and-errors. Thus, to avoid this, we focus on finding the optimal architecture that exactly meets the given latency requirement through a one-time search (i.e., \underline{\textit{you only search once}}).

\section{Methodology}
\label{sec:methodology}
In this section, we first elaborate on each component of the proposed LightNAS, and then discuss the relationships with previous methods to further distinguish the technical contributions of this work.

\begin{figure}[t]
    \begin{center}
        \includegraphics[width=0.9375\columnwidth]{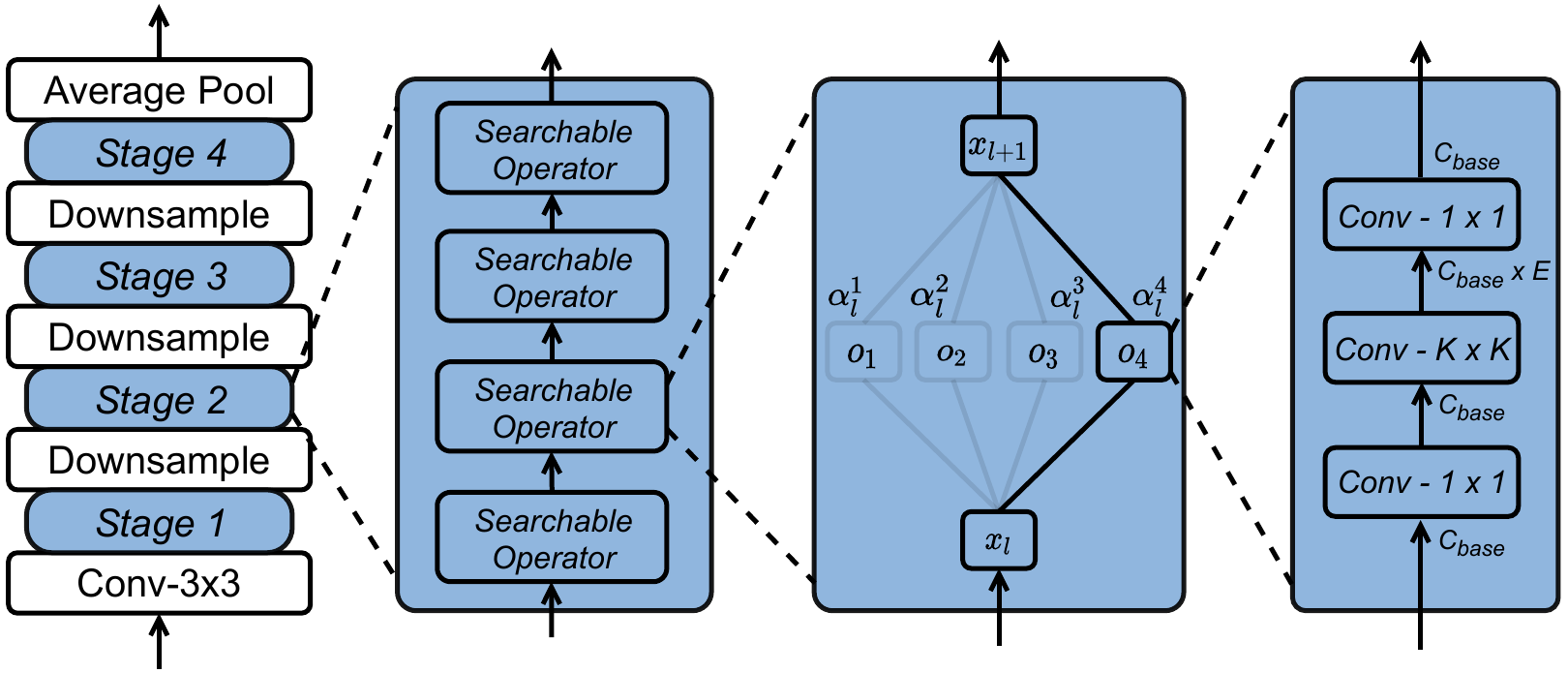}
    \end{center}
    \vspace{-3pt}
    \caption{Illustration of the supernet structure over the search space. \textit{K} and \textit{E} denote the kernel size and the expansion ratio, respectively.}
    \vspace{+3pt}
    \label{fig:search-space}
\end{figure}

\vspace{-4pt}
\subsection{Search Space Design}
\label{sec:search-space-design}

In the literature, differentiable NAS methods like DARTS \cite{liu2018darts} and its variants \cite{xie2018snas, kim2021mdarts} explicitly advocate for the cell-level architecture search. However, as pointed out in \cite{tan2019mnasnet}, enabling the layer diversity helps to strike the right balance between accuracy and efficiency. Thus, we closely follow the layer-wise architecture space design \cite{wu2019fbnet, cai2018proxylessnas, vahdat2020unas} as illustrated in Figure~\ref{fig:search-space}. Specifically, the operator space $\mathcal{O}$ is built upon MobileNetV2 \cite{sandler2018mobilenetv2}, in which we allow a set of \textit{MBConv} layers with diverse kernel sizes of $\{3,5,7\}$ and expansion ratios of $\{3,6\}$. Meanwhile, we include \textit{SkipConnect}, which is computation-free, to achieve flexible architecture search in terms of the network depth \cite{cai2018proxylessnas, wu2019fbnet}. As such, we have $|\mathcal{O}|=7$, and given that the supernet consists of $L=22$ searchable operators where the first one is fixed \cite{cai2018proxylessnas}, the architecture space size of LightNAS is then calculated as $|\mathcal{A}|=7^{21}\approx5.6\times10^{17}$. \underline{\textit{Unless specified otherwise}}, we do not apply extra techniques like Squeeze-and-Excitation (SE) module \cite{hu2018squeeze} and Swish activation \cite{howard2019searching} in order to ensure fair comparisons with previous hardware-aware NAS methods \cite{cai2018proxylessnas, wu2019fbnet, cai2019once, vahdat2020unas}.

\vspace{-4pt}
\subsection{Latency Prediction}
\label{sec:latency-prediction}
Nonetheless, given that the search space of NAS is prohibitively large (e.g., $|\mathcal{A}|\approx5.6\times10^{17}$ in LightNAS), measuring the on-device latency for every possible architecture $arch$ is computationally expensive \cite{cai2018proxylessnas}. To this end, we introduce an accurate yet efficient predictor to approximate the on-device latency for $arch \in \mathcal{A}$ with negligible computation overheads. With this goal in mind, we first encode $arch$ with a sparse matrix $\overline{\alpha} \in \{0,1\}^{L \times K}$, where $\overline{\alpha}_l^k=1$ indicates that the $k$-th operator is reserved for the $l$-th layer of $arch$ while others are discarded. As such, we can calculate $\overline{\alpha}$ as follows: 
\begin{equation}
    \small
    \setlength\abovedisplayskip{3pt}
    \setlength\belowdisplayskip{3pt}
    \overline{\alpha}_{l}^{k} = \begin{cases}
        1, \,\,\,\,\,\,\,\, \mathrm{if} \,\, \alpha_{l}^{k} \, = \, \mathrm{arg\,max} \, ||\alpha_{l}|| \\
        0, \,\,\,\,\,\,\,\, \mathrm{otherwise}
    \end{cases}
    \label{eq:alpha-sparse-matrix}
\end{equation}
Therefore, since the supernet is composed of $L$ searchable layers, we derive that the architecture encoding matrix $\overline{\alpha}$ contains $L$ entries with values of 1, whereas other entries are with values of 0.

\begin{figure}[t]
    \begin{center}
        \includegraphics[width=0.925\columnwidth]{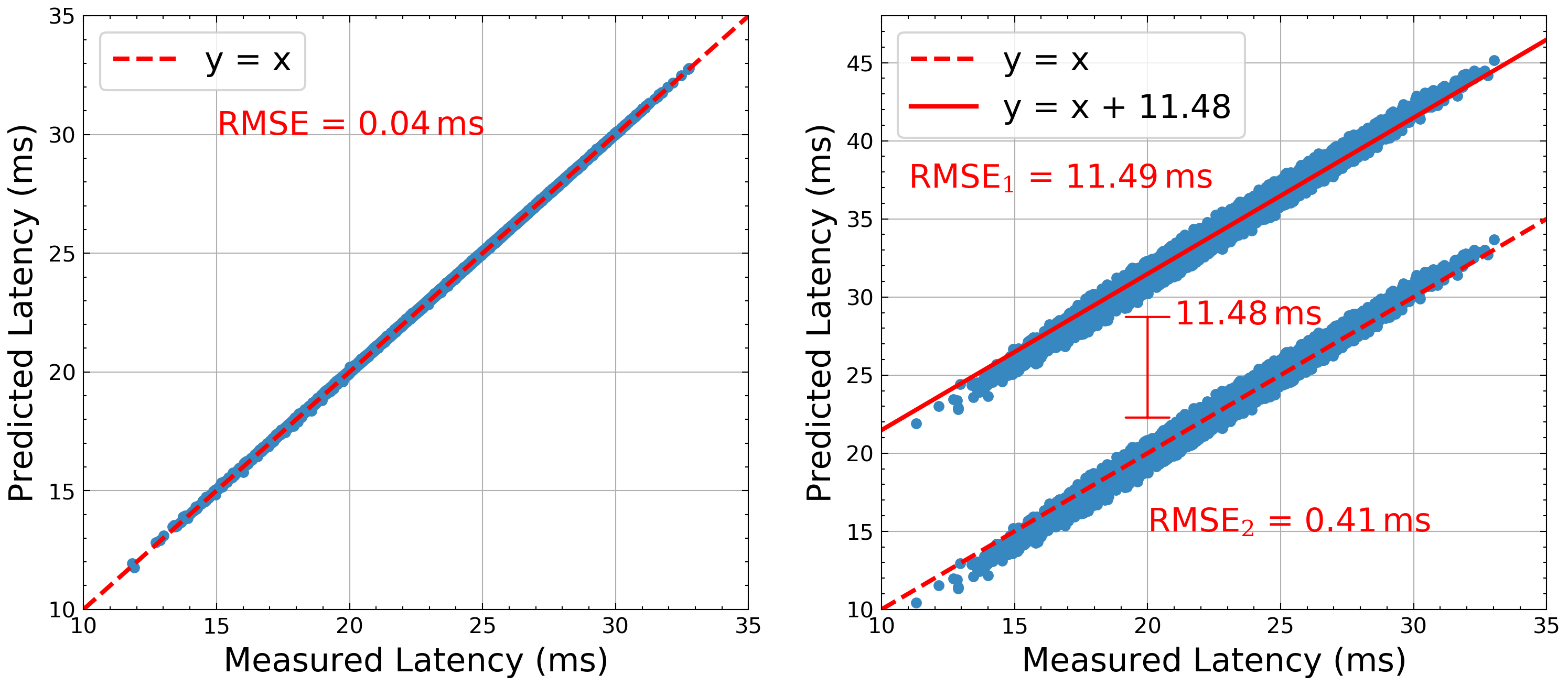}
    \end{center}
    \vspace{-3pt}
    \caption{\textit{Left}: the predicted results with the proposed latency predictor. \textit{Right}: the predicted results with the latency lookup table (LUT).}
    \label{fig:mlp-vs-lut}
\end{figure}

Subsequently, we leverage a multi-layer perceptron (MLP) model for the prediction purpose, which consists of three fully-connected layers with 128, 64, and 1 neurons. Meanwhile, we sample 10,000 random architectures from $\mathcal{A}$ and measure the inference latency on Nvidia Jetson AGX Xavier, respectively. The sampled architectures and latency measurements are then split into two folds with 80\% as the training set and 20\% as the validation set. Next, we flatten $\overline{\alpha}$ corresponding to each architecture and feed it into the MLP-based latency predictor. The predicted results on the validation set are illustrated in Figure~\ref{fig:mlp-vs-lut} (\textit{Left}), where we find that the proposed latency predictor achieves an extremely low root-mean-square error (RMSE) of 0.04\,ms. More importantly, once the latency predictor is well trained, it is able to estimate the on-device latency for $arch \in \mathcal{A}$ through a one-time inference, which takes less than one millisecond, and thus introduces trivial computation overheads.

Furthermore, we compare the proposed latency predictor with the latency lookup table (LUT) as widely used in recent NAS works \cite{cai2018proxylessnas, wu2019fbnet, cai2019once}. The predicted results of LUT are shown in Figure~\ref{fig:mlp-vs-lut} (\textit{Right}), where we find that there exists a consistent gap (about 11.48\,ms) between the predicted latency and the measured ground truth. And even though the above prediction gap is eliminated, the RMSE of LUT still remains 0.41\,ms, which is much worse than the proposed latency predictor. We note that the goal of this work is to search for the architecture that strictly meets the given latency requirement, and thus an accurate latency predictor is of great necessity.

\vspace{-4pt}
\subsection{Lightweight Architecture Search}
\label{sec:lightweight-differentiable-architecture-search}
Recall that previous differentiable NAS methods \cite{liu2018darts, xie2018snas, wu2019fbnet} require to simultaneously optimize multiple sub-networks (paths), inevitably causing the memory bottleneck \cite{cai2018proxylessnas} as well as violating the equality principle \cite{chu2021fairnas}. To this end, we propose a lightweight differentiable architecture search method to reduce the optimization complexity to the single-path level, thereby effectively resolving the memory bottleneck. Let $arch=\{op_l\}_{l=1}^{L}$ denote the stand-alone architecture candidate. Therefore, once the search process terminates, we can calculate the probability that $arch$ is selected as follows:
\begin{equation}
    \small
    \setlength\abovedisplayskip{3.25pt}
    \setlength\belowdisplayskip{3.25pt}
    P(arch) = \prod\nolimits_{l=1}^{L} P(op_l = o_k), \,\, s.t., \,\,o_k \in \mathcal{O}
    \label{eq:arch-probability}
\end{equation}
where $P(op_l=o_k)$ is the probability of $o_k$ being at the $l$-th layer:
\begin{equation}
    \small
    \setlength\abovedisplayskip{3.25pt}
    \setlength\belowdisplayskip{3pt}
    P(op_l = o_k) = \frac{\exp(\alpha_{l}^{k})}{\sum_{k'=1}^{K} \exp(\alpha_l^{k'})}
    \label{eq:op-probability}
\end{equation}
For the sake of simplicity, we replace $P(op_l=o_k)$ with $P_l^k$. To derive $arch$, one straightforward method is to optimize the architecture parameters $\alpha$ over the search space $\mathcal{A}$. However, given that the search space of LightNAS is prohibitively large as seen in Section~\ref{sec:search-space-design}, iterating every possible architecture over $\mathcal{A}$ inevitably requires a huge amount of computation resources \cite{liu2018darts, wu2019fbnet}. To alleviate this issue, we further leverage the Gumbel Softmax reparameterization trick \cite{jang2016categorical} to relax the discrete architecture space to be continuous:
\begin{equation}
    \small
    \setlength\abovedisplayskip{3.25pt}
    \setlength\belowdisplayskip{2.5pt}
    \widehat{P_{l}^{k}} = \frac{\exp[(P_{l}^{k} + G_{l}^{k})/\tau]}{\sum_{k'=1}^{K} \exp[(P_{l}^{k'} + G_{l}^{k'})/\tau]}
    \label{eq:gumbel-softmax}
\end{equation}
where $G \in \mathbb{R}^{L \times K}$ is the random variable drawn from $\mathrm{Gumbel(0,1)}$ \cite{jang2016categorical}. $\tau$ is the softmax temperature, which is initialized as 5 and then gradually decays to zero. We note that, once converged, the above relaxation is unbiased as proved in \cite{jang2016categorical} that $\mathrm{lim}_{\tau \to 0} \widehat{P}_l^k=P_l^k$. We then re-formulate the output of the supernet $F(x)$ as follows:
\begin{equation}
    \small
    \setlength\abovedisplayskip{3pt}
    \setlength\belowdisplayskip{2pt}
    F(x) = \sum_{l=1}^{L} \sum_{k=1}^{K} \left( \overline{P}_{l}^{k} \cdot o_k(x_l) \right), \,\, s.t., \,\, o_k \in \mathcal{O}
    \label{eq:gumbel-softmax-relax}
\end{equation}
where $\overline{P}$ is the binarization of $\widehat{P}$ that can be expressed as follows:
\begin{equation}
    \small
    \setlength\abovedisplayskip{3pt}
    \setlength\belowdisplayskip{2pt}
    \overline{P}_{l}^{k} = \begin{cases}
        1, \,\,\,\,\,\,\,\, \mathrm{if} \,\, \widehat{P}_{l}^{k} \, = \, \mathrm{arg\,max} \, ||\widehat{P}_{l}|| \\
        0, \,\,\,\,\,\,\,\, \mathrm{otherwise}
    \end{cases}
    \label{eq:p-sparse-matrix}
\end{equation}
As a result, we only need to activate one single-path sub-network during the forward propagation of the supernet since $\overline{P} \in \{0,1\}^{L \times K}$. The intuition behind this is that the output of the supernet only depends on operators with $\overline{P}_l^k=1$ as seen in Eq~(\ref{eq:gumbel-softmax-relax}). To summarize, the above single-path mechanism achieves two main benefits. On the one hand, it brings significant memory efficiency because the optimization complexity has been reduced to the single-path level, and considering that the GPU memory is constant, we are allowed to use a larger batch size to speed up the search process. On the other hand, the above single-path mechanism forces the search process to strictly satisfy the equality principle \cite{chu2021fairnas}, i.e., the supernet and the searched sub-network should be trained in the same manner.

\vspace{-4pt}
\subsection{Hardware-Aware Architecture Search}
\label{sec:hardware-aware-differentiable-architecture-search}

The search method described above merely optimizes the search process in terms of the accuracy, while ignoring other critical performance metrics, e.g., the on-device latency as the most dominant one \cite{tan2019mnasnet}. Thus, we further integrate the latency predictor into LightNAS to achieve hardware-aware architecture search. The optimization objective of LightNAS is then formulated as follows:
\begin{equation}
    \small
    \setlength\abovedisplayskip{3pt}
    \setlength\belowdisplayskip{2pt}
    \mathop{\mathrm{minimize}}_{\alpha} \,\, \mathcal{L}_{valid}(w^*(\alpha), \alpha) + \lambda \cdot \left( \frac{LAT(\alpha)}{T} - 1\right)
    \label{eq:objective-lightnas}
\end{equation}
where $LAT(\alpha)$ represents the predicted latency of the architecture encoded by $\alpha$, and $T$ is the specified latency constraint. Besides, $\lambda$ denotes the coefficient to control the trade-off magnitude between accuracy and latency. Different from previous NAS methods \cite{wu2019fbnet, cai2018proxylessnas}, $\lambda$ in Eq~(\ref{eq:objective-lightnas}) is not a constant but a hyper-parameter to be optimized. Therefore, instead of manual hyper-parameter tuning, LightNAS automatically learns the optimal hyper-parameter configuration during the search process, which maximizes the accuracy while strictly satisfying the specified latency constraint $LAT(\alpha)=T$. For the sake of simplicity, we use $\mathcal{L}(w, \alpha, \lambda)$ to denote the objective defined in Eq~(\ref{eq:objective-lightnas}). Subsequently, $w$ and $\alpha$ are updated with gradient descent \cite{liu2018darts}, whereas $\lambda$ is optimized using \underline{\textit{gradient ascent}}: 
\begin{equation}
    \small
    \setlength\abovedisplayskip{3pt}
    \setlength\belowdisplayskip{2pt}
    \begin{cases}
        w^* = w - \eta_w \cdot \frac{\partial\mathcal{L}(w, \alpha, \lambda)}{\partial w}, \,\,
        \alpha^* = \alpha - \eta_{\alpha} \cdot \frac{\partial \mathcal{L}(w, \alpha, \lambda)}{\partial \alpha} \\ 
        \lambda^* = \lambda + \eta_{\lambda} \cdot \frac{\partial \mathcal{L}(w, \alpha, \lambda)}{\partial \lambda} = \lambda + \eta_{\lambda} \cdot \left(\frac{LAT(\alpha)}{T} - 1\right)
    \end{cases}
    \label{eq:parameter-update}
\end{equation}
where $\eta_{w}$, $\eta_{\alpha}$, and $\eta_{\lambda}$ are the learning rates of $w$, $\alpha$, and $\lambda$, respectively. After demonstrating \textit{what} the proposed method is, we then analyze \textit{why} it guarantees $LAT(\alpha)=T$. Recall that a larger $\lambda$ derives the architecture with low latency, whereas a smaller $\lambda$ generates the architecture with high latency as shown in Figure~\ref{fig:tradeoff}. Thus, if $LAT(\alpha)>T$, the gradient ascent scheme increases $\lambda$ to reinforce the latency regularization magnitude. As a result, $LAT(\alpha)$ decreases towards $T$ in the next search iteration. Likewise, if $LAT(\alpha)<T$, the gradient ascent scheme then decreases $\lambda$ to diminish the latency regularization magnitude, and the search engine therefore increases $LAT(\alpha)$ towards $T$ in the next parameter update. Consequently, the search engine ends up with the architecture with optimal accuracy while at the same time satisfying the given constraint $LAT(\alpha)=T$. To summarize, unlike previous hardware-aware differentiable NAS methods \cite{cai2018proxylessnas, wu2019fbnet, vahdat2020unas} that require multiple trial-and-errors to find the desired architecture with the latency of $T$, LightNAS only needs to search once, greatly improving the search efficiency.

\begin{table}[t]
\centering
\resizebox{1.0\linewidth}{!}{
    \begin{tabular}{lc|c|c|c|c|c|c|}
        & \rot{\small \cite{liu2018darts}}
        & \rot{\small \cite{tan2019mnasnet}}
        & \rot{\small \cite{cai2019once}}
        & \rot{\small \cite{wu2019fbnet}}
        & \rot{\small \cite{cai2018proxylessnas}}
        & \rot{\small Ours}\\
         \multicolumn{1}{l|}{Differentiable} & \cmark & \xmark & \xmark & \cmark & \cmark & \cmark \\
         \multicolumn{1}{l|}{Latency Optimization} & \xmark & \cmark & \cmark &  \cmark &  \cmark &  \cmark \\
         \multicolumn{1}{l|}{Specified Latency} & \xmark & \cmark & \cmark & \xmark & \xmark  & \cmark  \\
         \multicolumn{1}{l|}{Proxyless Search} & \xmark & \cmark & \cmark & \cmark & \cmark  & \cmark   \\
         \multicolumn{1}{l|}{Search Complexity} & $\mathcal{O}(K^2)$ & $\mathcal{O}(1)$ & $\mathcal{O}(1)$ & $\mathcal{O}(K^2)$ & $\mathcal{O}(2^2)$ & $\mathcal{O}(1)$ \\
         \multicolumn{1}{l|}{Search Cost (GPU hours)} & 24 & 40,000 & 1,275 & 216 & 200 & 10 \\
    \end{tabular}
}
\vspace{+2pt}
\caption{Comparisons with previous state-of-the-art NAS approaches.}
\label{tab:comparisons}
\end{table}

Meanwhile, given that $\mathcal{L}(w, \alpha, \lambda)$ is differentiable with respect to $w$ as seen in \cite{liu2018darts}, we then provide the differentiable analysis of $\alpha$:
\begin{equation}
    \small
    \setlength\abovedisplayskip{3pt}
    \setlength\belowdisplayskip{2pt}
    \begin{aligned}
        &\frac{\partial \mathcal{L}(w, \alpha, \lambda)}{\partial \alpha} = \frac{\partial \mathcal{L}_{valid}}{\partial \alpha} + \frac{\lambda}{T} \cdot \frac{\partial LAT(\alpha)}{\partial \alpha}  \\
        &= \frac{\partial \mathcal{L}_{valid}}{\partial \overline{P}} \cdot \frac{\partial \overline{P}}{\partial \widehat{P}} \cdot \frac{\partial \widehat{P}}{\partial P} \cdot \frac{\partial P}{\partial \alpha} + \frac{\lambda}{T} \cdot \frac{\partial LAT(\alpha)}{\partial \overline{P}} \cdot \frac{\partial \overline{P}}{\partial \widehat{P}} \cdot \frac{\partial \widehat{P}}{\partial P} \cdot \frac{\partial P}{\partial \alpha}
    \end{aligned}
    \label{eq:differentiable-analysis}
\end{equation}
where $\frac{\partial \overline{P}}{\partial \widehat{P}} \approx \mathbb{1}$ because $\overline{P}$ is the binarization of $\widehat{P}$ \cite{bengio2013estimating}. With the equivalence of $\overline{\alpha}$ and $\overline{P}$ as seen in Eq~(\ref{eq:alpha-sparse-matrix}) and Eq~(\ref{eq:p-sparse-matrix}), $\frac{\partial LAT(\alpha)}{\partial \overline{P}}$ is determined by the network weights of the latency predictor, which can be obtained through a one-time backward propagation. In addition to these terms, other terms in Eq~(\ref{eq:differentiable-analysis}) are apparently differentiable since only continuous transformations are involved \cite{liu2018darts, jang2016categorical}. Please note that the differentiable analysis of $\lambda$ is given in Eq~(\ref{eq:parameter-update}).

\vspace{-4pt}
\subsection{Relationships with Previous Methods}
\label{sec:discussions-and-relationship-to-previous-works}

First of all, to obtain the architecture that satisfies the given latency constraint, previous hardware-aware differentiable NAS methods \cite{cai2018proxylessnas, wu2019fbnet, vahdat2020unas} require to perform a hyper-parameter sweep to manually tune the trade-off coefficient $\lambda$ by trial and error (see Section~\ref{sec:motivations}). The total design cost therefore increases proportionally (empirically by $\times10$ times). We note that previous NAS methods only report the explicit search cost such as the time needed to run one single search, whereas the implicit search cost like the time required to manually tune $\lambda$ is excluded. In contrast, this paper focuses on finding the required architecture through a one-time search so as to eliminate the implicit search cost, thereby bringing considerable search efficiency and flexibility. It is worth noting that reinforcement learning and evolution-based NAS approaches \cite{tan2019mnasnet, cai2019once} can achieve the same goal as LightNAS, but suffer from prohibitive search overheads (e.g., 40,000 GPU hours in \cite{tan2019mnasnet} and 10 GPU hours in LightNAS).

Meanwhile, due to the multi-path paradigm, previous differentiable NAS methods \cite{liu2018darts, xie2018snas, wu2019fbnet} suffer from the memory bottleneck, which violate the equality principle as well \cite{chu2021fairnas}. To this end, we propose the lightweight differentiable architecture search method, effectively reducing the optimization complexity to the single-path level. Owing to the single-path mechanism, the optimization gap between the supernet and the searched sub-network is bridged \cite{chu2021fairnas}. Besides, different from the latency lookup table (LUT) \cite{wu2019fbnet}, the proposed predictor can approximate not only the on-device latency but also other hardware metrics like the runtime energy as seen in Figure~\ref{fig:energy-search}. Without loss of generality, LightNAS can be effortlessly plugged into various scenarios, in which we only need to replace the latency predictor with the predictor of the target scenario. Finally, we compare LightNAS against previous NAS methods in Table~\ref{tab:comparisons}.
\begin{figure}[t]
    \begin{center}
        \includegraphics[width=1.0\columnwidth]{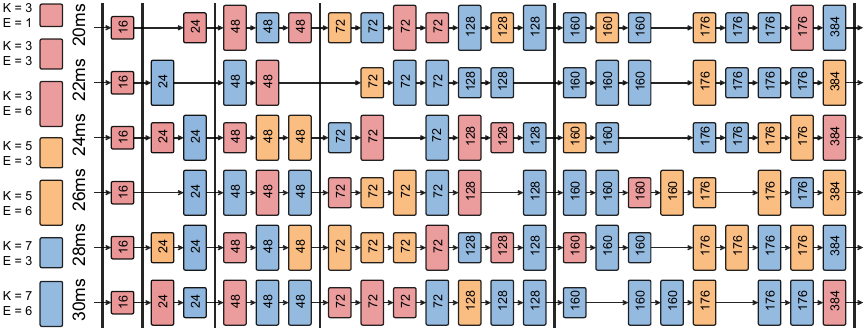}
    \end{center}
    \vspace{-3.5pt}
    \caption{Illustration of LightNets under different latency constraints. The integer in each operator denotes the number of base channels.}
    \vspace{+3pt}
    \label{fig:architectures}
\end{figure}

\begin{figure}[t]
    \begin{center}
        \includegraphics[width=0.98\columnwidth]{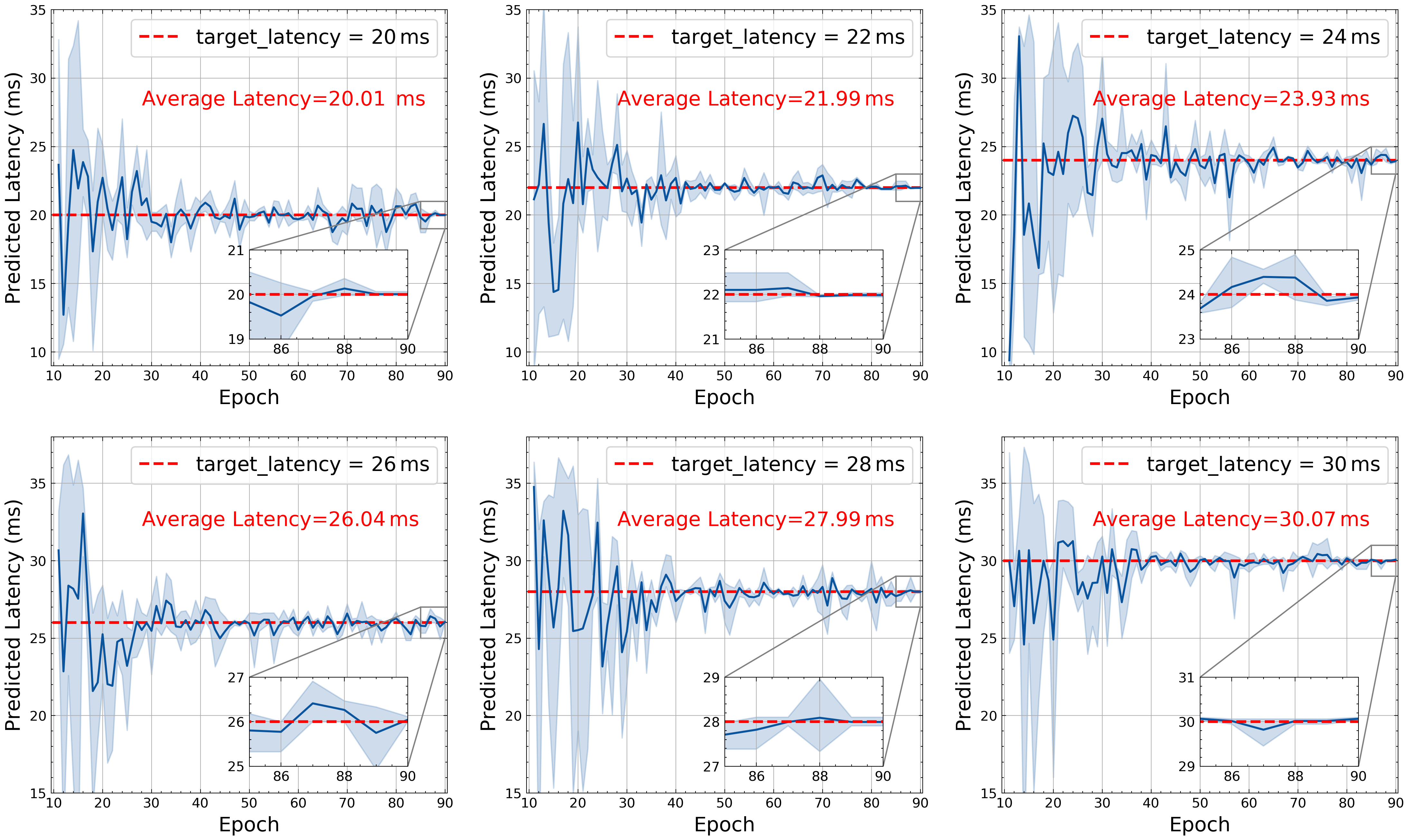}
    \end{center}
    \vspace{-3.5pt}
    \caption{Illustration of the search process under various constraints.}
    \label{fig:latency-search}
    \vspace{+3pt}
\end{figure}

\section{Experiments}
\label{sec:experiments}
In this section, we extensively evaluate the proposed LightNAS on a cutting-edge embedded platform called Nvidia Jetson AGX Xavier. Here, the MAXN power mode is applied to maximize the hardware performance. Meanwhile, to avoid resource underutilization, all the measurements are reported with an input batch size of 8 \cite{cai2018proxylessnas}.

\vspace{-4pt}
\subsection{Dataset and Implementation Details}
\label{sec:dataset-and-implementation-details}

\textbf{\textit{Dataset.}} 
All the experiments are directly conducted on ImageNet \cite{deng2009imagenet}. Specifically, ImageNet consists of 1,000 categories, and 1.28M training images and 50K validation images, all of which are roughly equally distributed across all categories. Following the network design conventions \cite{sandler2018mobilenetv2, tan2019mnasnet}, we use the mobile setting, where the image size is set to $224\times224$, and the number of multi-add operations is strictly under 600M during the runtime inference.

\noindent
\textbf{\textit{Architecture Search Settings.}} 
In LightNAS, the architecture search settings closely follow FBNet \cite{wu2019fbnet}, in which we randomly sample 100 categories from ImageNet to optimize both network weights $w$ and architecture parameters $\alpha$. Specifically, we train the stochastic supernet for 90 epochs with an input batch size of 128. In the first 10 epochs, we only update $w$, whereas $\alpha$ is frozen \cite{wu2019fbnet}. Subsequently, the optimization steps of $w$ and $\alpha$ alternate in each epoch. To optimize $w$, we use the SGD optimizer with a learning rate of 0.1 (annealed down to zero following the cosine schedule), a momentum of 0.9, and a weight decay of $3\times10^{-5}$. To optimize $\alpha$, we employ the Adam optimizer \cite{liu2018darts} with a learning rate of 0.001 and a weight decay of $1\times10^{-3}$. Besides, as discussed in Section~\ref{sec:hardware-aware-differentiable-architecture-search}, the trade-off coefficient $\lambda$ in LightNAS is not a constant but a parameter to be optimized. For this reason, we initialize $\lambda$ as zero and optimize $\lambda$ with the gradient ascent scheme, where the learning rate is fixed to 0.0005. Finally, we denote the architectures searched by LightNAS as LightNets. It is worth noting that all the architecture search experiments are conducted on one single GeForce RTX 3090 GPU. 

\begin{table}[t]
\centering
\resizebox{1.0\linewidth}{!}{%
\scriptsize
\begin{tabular}{l|c|c|c|c|c}
\toprule[0.15em]
\multirow{2}{*}{Architecture} & \multirow{2}{*}{Method} & Search Cost & \multicolumn{2}{c|}{Accuracy (\%)} & Latency \\ \cline{4-5}
 &  & (GPU hours) & Top-1 & Top-5 & (ms) \\ \hline
MobileNetV2 \cite{sandler2018mobilenetv2} & Manual & - & 72.0 & 91.0 & 20.2 \\ 
ProxylessNAS \cite{cai2018proxylessnas} & Differentiable & 200 & 74.6 & 92.2 & 21.2 \\ FBNet-A \cite{wu2019fbnet} & Differentiable & 216 & 73.0 & 90.9 & 21.7 \\
OFA-S \cite{cai2019once} & Evolution & 1,275 & 72.9 & 91.1 & 21.4 \\
MnasNet-B1 \cite{tan2019mnasnet} & Reinforcement & 40,000 & 74.5 & 92.1 & 20.1 \\ 
\rowcolor{Gray} LightNet-20ms& Differentiable & 10 & 75.0 & 92.2 & 20.0 \\ \hline
FBNet-B \cite{wu2019fbnet} & Differentiable & 216 & 74.1 & 91.8 & 23.0 \\
MobileNetV3$^{\dagger}$ \cite{howard2019searching} & Manual & - & 75.2 & - & 23.0 \\
MnasNet-A1$^{\dagger}$ \cite{tan2019mnasnet} & Reinforcement & 40,000 & 75.2 & 92.5 & 22.9 \\
\rowcolor{Gray} LightNet-22ms& Differentiable & 10 & 75.2 & 92.2 & 22.1 \\ \hline
ProxylessNAS \cite{cai2018proxylessnas} & Differentiable & 200 & 75.1 & 92.5 & 24.5 \\ UNAS \cite{vahdat2020unas} & Differentiable & 103 & 75.3 & 92.4 & 24.2 \\
FBNet-Xavier \cite{wu2019fbnet} & Differentiable & $\sim186$ & 74.6 & 92.1 & 24.1 \\ 
\rowcolor{Gray} LightNet-24ms& Differentiable & 10 & 75.5 & 92.3 & 23.9 \\ \hline
FBNet-C \cite{wu2019fbnet} & Differentiable & 216 & 74.9 & 92.3 & 26.4 \\
OFA-M \cite{cai2019once} & Evolution & 1,275 & 75.4 & 92.4 & 26.3 \\ 
\rowcolor{Gray} LightNet-26ms& Differentiable & 10 & 75.9 & 92.6 & 26.1 \\ \hline
OFA-L \cite{cai2019once} & Evolution & 1,275 & 75.8 & 92.7 & 29.3 \\ 
ProxylessNAS \cite{cai2018proxylessnas} & Differentiable & 200 & 75.3 & - & 29.9 \\
\rowcolor{Gray} LightNet-28ms& Differentiable & 10 & 76.1 & 92.7 & 28.2 \\ \hline
EfficientNet-B0$^{\dagger}$ \cite{tan2019efficientnet} & Reinforcement & - & 76.3 & - & 37.2 \\ 
\rowcolor{Gray} LightNet-30ms& Differentiable & 10 & 76.4 & 92.9 & 30.1 \\ 
\toprule[0.15em]
\end{tabular}%
}
\vspace{+2pt}
\caption{Comparisons with state-of-the-art architectures on ImageNet \cite{deng2009imagenet}. $\dagger$ denotes architectures that use extra techniques like Swish activation and Squeeze-and-Excitation (SE) module \cite{hu2018squeeze, howard2019searching}.}
\label{tab:imagenet}
\end{table}

\noindent
\textbf{\textit{Architecture Evaluation Settings.}} 
We simply follow the training protocols as widely used in previous NAS methods \cite{cai2018proxylessnas, wu2019fbnet} to evaluate the searched LightNets on ImageNet \cite{deng2009imagenet}. Specifically, we retrain LightNets from scratch for 360 epochs with a batch size of 1024 on 4 GeForce RTX 3090 GPUs, in which the standard data augmentations are applied \cite{wu2019fbnet} throughout this work. The default optimizer is SGD with a momentum of 0.9 and a weight decay of $4\times10^{-5}$. Besides, the learning rate is initialized as 0.5, which gradually decays to zero following the cosine schedule. Similar to DARTS \cite{liu2018darts}, we linearly warm up the learning rate from 0.1 to 0.5 in the first 5 epochs. Meanwhile, we insert the Dropout module before the final classification layer, where the dropout ratio is set to 0.2 \cite{wu2019fbnet}.

\vspace{-5pt}
\subsection{Experimental Results}
\label{sec:experimental-results}

\noindent
\textbf{\textit{Architecture Search Results.}} 
We visualize the LightNets searched under different latency constraints in Figure~\ref{fig:architectures}, which span from 20\,ms to 30\,ms. Different from MobileNetV2 \cite{sandler2018mobilenetv2} that simply stacks the same operator across all network layers, LightNAS effectively enables the layer diversity to strike the right balance between accuracy and latency. Meanwhile, given a larger latency constraint, we observe that the search engine of LightNAS encourages to search for the desired architecture that goes deeper and wider. 

\noindent
\textbf{\textit{Architecture Evaluation Results.}} 
Results and comparisons with previous state-of-the-art architectures are summarized in Table~\ref{tab:imagenet}, where we find that the searched LightNets strictly satisfy the given latency constraints, while at the same time coming at an extremely low search cost of 10 GPU hours. More importantly, all LightNets are obtained through a one-time search, and thus the manual trial-and-errors over the trade-off coefficient $\lambda$ are eliminated (see Section~\ref{sec:motivations}). Meanwhile, under the same latency constraints, LightNets consistently outperform previous state-of-the-art architectures in terms of the accuracy on ImageNet \cite{deng2009imagenet}. We note that FBNet \cite{wu2019fbnet} is the most relevant work to LightNAS in the literature. For comparisons, we implement FBNet and exploit FBNet to perform the architecture search on Nvidia Jetson AGX Xavier, where the searched architecture is denoted as FBNet-Xavier. As shown in Table~\ref{tab:imagenet}, LightNet-24ms achieves a +0.9\% higher top-1 accuracy than FBNet-Xavier while with a comparable latency constraint of 24\,ms.

\begin{figure}[t]
    \begin{center}
        \includegraphics[width=1.0\columnwidth]{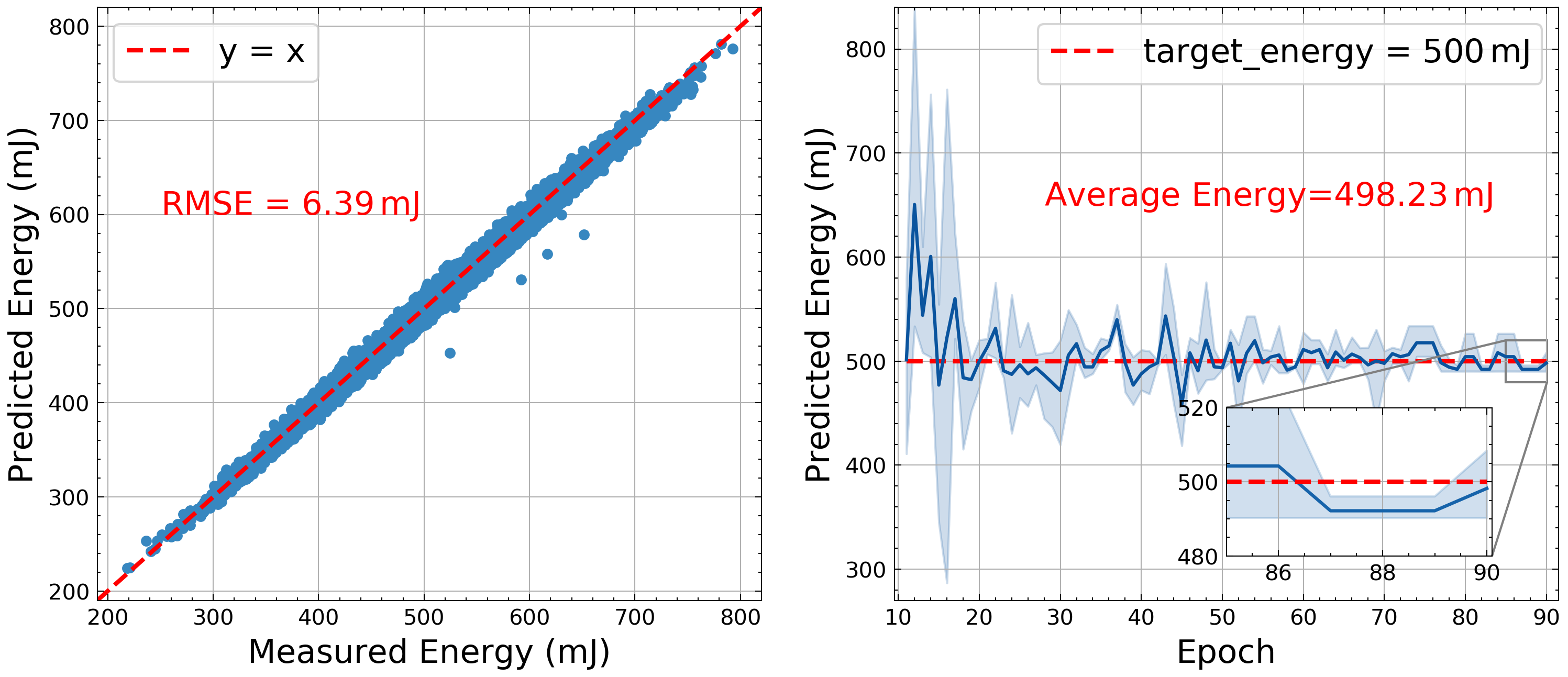}
    \end{center}
    \vspace{-3pt}
    \caption{\textit{Left}: the predicted results with the predictor in Section \ref{sec:latency-prediction}. \textit{Right}: the search process under the energy constraint of 500\,mJ.}
    \vspace{+3pt}
    \label{fig:energy-search}
\end{figure}

\begin{figure}[t]
    \begin{center}
        \includegraphics[width=1.0\columnwidth]{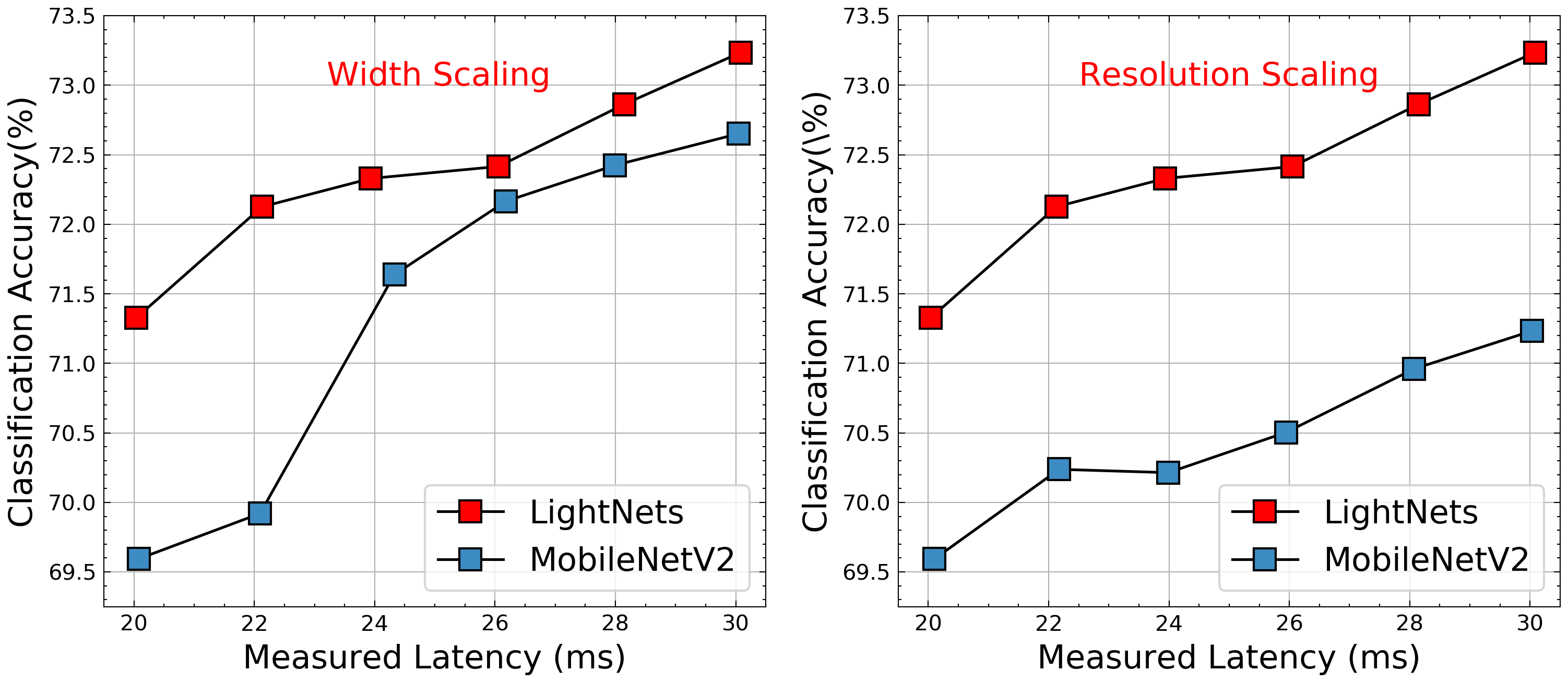}
    \end{center}
    \vspace{-3pt}
    \caption{Performance comparisons with different model scaling techniques \cite{tan2019mnasnet}. Here, all the models are trained for only 50 epochs.}
    \vspace{+3pt}
    \label{fig:mobilenetv2-width-res-scaling}
\end{figure}

\vspace{-4pt}
\subsection{Ablation Studies and Discussions}
\label{sec:ablation-studies-and-analysis}

\noindent
\textbf{\textit{Architecture Search Stability.}} 
We visualize the search process of LightNAS under various latency constraints in Figure~\ref{fig:latency-search}, where each figure is drawn by averaging three different search runs. Notably, LightNAS always ends up with the architecture that strictly meets the given latency constraint. Meanwhile, the search engine explicitly focuses on searching for the optimal architecture around the target latency, which aligns with the analysis in Section~\ref{sec:hardware-aware-differentiable-architecture-search}.
 
\noindent
\textbf{\textit{Generality to Energy-Critical Tasks.}}
To generalize LightNAS to the energy-critical scenarios, we first apply the proposed predictor to approximate the energy consumption for $arch \in \mathcal{A}$ as shown in Figure~\ref{fig:energy-search} (\textit{Left}). Please note that the energy measurement inevitably suffers from noises caused by the hardware temperature. Then, we integrate the energy predictor into LightNAS and visualize the search process in Figure~\ref{fig:energy-search} (\textit{Right}), in which find that LightNAS is able to effectively generalize to the energy-critical tasks. 

\noindent
\textbf{\textit{Transferability to Object Detection.}}
We next evaluate LightNets on the object detection task, in which we use a popular detection framework named SSDLite \cite{liu2016ssd} and treat different architectures as drop-in backbone replacements. All the architectures are trained from scratch (i.e., without loading pretrained weights) under the same settings on COCO2017. As summarized in Table~\ref{tab:object-detection}, LightNets achieve better performance than the compared architectures in terms of both detection accuracy and execution efficiency.

\noindent
\textbf{\textit{Comparisons with Scaling Techniques.}}
Another alternative to guarantee the specified latency requirements is the model scaling technique \cite{tan2019mnasnet}. Given that the search space of LightNAS is based on MobileNetV2 \cite{sandler2018mobilenetv2}, we further scale up MobileNetV2 with respect to width/resolution to accommodate different latency requirements. As seen in Figure~\ref{fig:mobilenetv2-width-res-scaling}, under the same latency constraints, LightNets clearly outperform those counterparts in terms of the accuracy.

\noindent
\textbf{\textit{Ablation of Squeeze-and-Excitation Module.}}
Previous methods \cite{tan2019mnasnet, howard2019searching, tan2019efficientnet} use extra techniques like Squeeze-and-Excitation (SE) \cite{hu2018squeeze} to improve the performance as shown in Table~\ref{tab:imagenet}. Therefore, for comparisons, we apply the SE module to the last nine layers of LightNets. As seen in Table~\ref{tab:effect-of-se}, the SE module greatly improves the accuracy of LightNets while slightly sacrificing the efficiency.

\begin{table}[t]
\centering
\resizebox{1.0\linewidth}{!}{%
\scriptsize
\begin{tabular}{l|c|c|c|c|c|c|c}
\toprule[0.15em]
Backbone & $\mathrm{AP}$ & $\mathrm{AP_{50}}$ & $\mathrm{AP_{75}} $ & $\mathrm{AP_S}$ & $\mathrm{AP_M}$ & $\mathrm{AP_L}$ & Latency (ms) \\ \hline
ProxylessNAS \cite{cai2018proxylessnas} & 20.3 & 34.6 & 20.3 & 2.2 & 19.3 & 39.6 & 70.1 \\ 
MobileNetV2 \cite{sandler2018mobilenetv2} & 20.4 & 34.3 & 20.5 & 1.6 & 19.5 & 40.2 & 72.6 \\ 
MnasNet-A1 \cite{tan2019mnasnet} & 21.2 & 36.0 & 21.4 & 2.5 & 20.6 &41.5 & 74.2  \\
FBNet-C \cite{wu2019fbnet} & 21.5 & 36.2 & 21.9 & 2.5 & 20.9 & 41.5 & 76.5 \\ 
OFA-M \cite{cai2019once} & 21.6 & 36.7 & 21.9 & 2.2 & 21.4 & 41.3 & 75.4  \\  \hline
\rowcolor{Gray} LightNet-20ms & 20.8 & 35.2 & 21.2 & 1.9 & 19.9 & 41.0 & 67.1 \\ 
\rowcolor{Gray} LightNet-24ms & 21.5 & 36.3 & 21.7 & 2.5 & 21.2 & 42.2 & 68.6 \\ 
\rowcolor{Gray} LightNet-28ms & 21.9 & 36.9 & 22.0 & 2.4 & 21.9 & 41.8 & 69.7 \\ 
\toprule[0.15em]
\end{tabular}%
}
\vspace{-3pt}
\caption{Comparisons with state-of-the-art backbones on COCO2017.}
\vspace{+1pt}
\label{tab:object-detection}
\end{table}

\begin{table}[t]
\centering
\resizebox{1.0\linewidth}{!}{%
\scriptsize
\begin{tabular}{c|c|c|c|c|c}
\toprule[0.15em]
\multirow{2}{*}{} & \multirow{2}{*}{Architecture} & \multicolumn{2}{c|}{Accuracy (\%)} &  \multirow{2}{*}{FLOPs (M)} & \multirow{2}{*}{Latency (ms)} \\ \cline{3-4}
 &  & \multicolumn{1}{c|}{Top-1} & \multicolumn{1}{c|}{Top-5} &  \\ \hline
\multirow{6}{*}{\rotatebox{90}{With SE \cite{hu2018squeeze}}} 
 & LightNet-20ms-SE & 75.4 \textcolor{secolor}{(+0.4)} & 92.3 \textcolor{secolor}{(+0.1)} & 356 \textcolor{secolor}{(+2)} & 20.9 \textcolor{secolor}{(+0.9)} \\ \cline{2-6} 
 & LightNet-22ms-SE & 76.1 \textcolor{secolor}{(+0.9)} & 92.5 \textcolor{secolor}{(+0.3)} & 352 \textcolor{secolor}{(+3)} & 23.2 \textcolor{secolor}{(+1.1)} \\ \cline{2-6} 
 & LightNet-24ms-SE & 75.9 \textcolor{secolor}{(+0.4)} & 92.6 \textcolor{secolor}{(+0.3)} & 385 \textcolor{secolor}{(+2)} & 25.5 \textcolor{secolor}{(+1.6)} \\ \cline{2-6} 
 & LightNet-26ms-SE & 76.3 \textcolor{secolor}{(+0.4)} & 92.8 \textcolor{secolor}{(+0.2)} & 435 \textcolor{secolor}{(+3)} & 27.7 \textcolor{secolor}{(+1.6)} \\ \cline{2-6} 
 & LightNet-28ms-SE & 76.5 \textcolor{secolor}{(+0.4)} & 92.8 \textcolor{secolor}{(+0.1)} & 464 \textcolor{secolor}{(+4)} & 30.3 \textcolor{secolor}{(+2.1)} \\ \cline{2-6} 
 & LightNet-30ms-SE & 77.0 \textcolor{secolor}{(+0.6)} & 93.1 \textcolor{secolor}{(+0.2)} & 493 \textcolor{secolor}{(+4)} & 31.9 \textcolor{secolor}{(+1.8)} \\ 
\toprule[0.15em]
\end{tabular}%
}
\vspace{-3pt}
\caption{Ablation of the Squeeze-and-Excitation (SE) module \cite{hu2018squeeze}.}
\vspace{-2pt}
\label{tab:effect-of-se}
\end{table}

\section{Conclusion}
\label{sec:conclusion}
This paper proposes, designs, and validates a lightweight hardware-aware differentiable NAS framework dubbed LightNAS. In contrast to previous NAS methods that require a plethora of trial-and-errors, LightNAS is able to effectively and efficiently find the architecture that strictly satisfies the specified performance constraint through a one-time search. Extensive experiments are conducted to show the superiority of LightNAS over state-of-the-art NAS approaches.

\balance

\vspace{-3pt}
\bibliographystyle{unsrt}
\bibliography{main}

\end{document}